\documentclass{article} % For LaTeX2e
\usepackage{mlgenx2024,times}

\usepackage{amsmath,amsfonts,bm}

\def\eqref#1{equation~\ref{#1}}
\def\1{\bm{1}}

\newcommand{\R}{\mathbb{R}}
\newcommand{\E}{\mathbb{E}}

\newcommand{\N}{\mathcal{N}}

\newcommand{\I}{\mathbf{I}}
\newcommand{\B}{\mathbf{B}}

\newcommand{\W}{\mathbf{W}}
\newcommand{\K}{\mathbf{K}}
\newcommand{\X}{\mathbf{X}}
\newcommand{\U}{\mathbf{U}}
\newcommand{\Y}{\mathbf{Y}}
\newcommand{\Z}{\mathbf{Z}}
\newcommand{\cL}{\mathcal{L}}

\newcommand{\m}{\mathbf{m}}
\newcommand{\bU}{\mathbf{U}}
\newcommand{\F}{\mathbf{F}}

\newcommand{\x}{\mathbf{x}}
\newcommand{\y}{\mathbf{y}}
\newcommand{\f}{\mathbf{f}}
\newcommand{\bu}{\mathbf{u}}
\newcommand{\0}{\bm{0}}
\newcommand{\KL}{\text{KL}}

\DeclareMathAlphabet{\mathsfit}{\encodingdefault}{\sfdefault}{m}{sl}
\SetMathAlphabet{\mathsfit}{bold}{\encodingdefault}{\sfdefault}{bx}{n}

\usepackage{hyperref}
\usepackage{url}

\usepackage{graphicx}

\usepackage{subcaption}
\usepackage{cleveref}
\usepackage{colortbl} % Required for coloring the columns

\definecolor{pastelgreen}{RGB}{182, 217, 193}
\definecolor{pastelblue}{RGB}{182, 205, 217}

\title{Scalable Amortized GPLVMs for Single Cell Transcriptomics Data}

\author{Sarah Zhao  \\
Department of Statistics\\
Stanford University\\
Stanford, CA 94305, USA \\
\texttt{smxzhao@stanford.edu} \\
\And Aditya Ravuri \\
Department of Computer Science\\
University of Cambridge\\
Cambridge, United Kingdom\\
\texttt{ar847@cam.ac.uk} \\
\AND
Vidhi Lalchand \\
Eric and Wendy Schmidt Center\\
Broad Institute of MIT and Harvard \\
Cambridge, MA 02142, USA \\
\texttt{vidrl@mit.edu } \\
\And
Neil D. Lawrence \\
Department of Computer Science\\
University of Cambridge\\
Cambridge, United Kingdom\\
\texttt{ndl21@cam.ac.uk}\\
}

\iclrfinalcopy % Uncomment for camera-ready version, but NOT for submission.
\begin{document}

\maketitle

\begin{abstract}
Dimensionality reduction is crucial for analyzing large-scale single-cell RNA-seq data. Gaussian Process Latent Variable Models (GPLVMs) offer an interpretable dimensionality reduction method, but current scalable models lack effectiveness in clustering cell types. We introduce an improved model, the amortized stochastic variational Bayesian GPLVM (BGPLVM), tailored for single-cell RNA-seq with specialized encoder, kernel, and likelihood designs. This model matches the performance of the leading single-cell variational inference (scVI) approach on synthetic and real-world COVID datasets and effectively incorporates cell-cycle and batch information to reveal more interpretable latent structures as we demonstrate on an innate immunity dataset.
\end{abstract}

\section{Introduction}
Single-cell transcriptomics sequencing (scRNA-seq) has enabled the study of gene expression at the individual cell level.
% unveiling cell-gene dependencies at unprecedented resolutions unobtainable by bulk experiments \citep{luecken2019currentsctutorial_survey, hie2020computationalmethods_survey}. 
This high-resolution analysis has helped discover new cell types and cell states, reveal developmental lineages, and identify cell type-specific gene expression profiles \citep{montoro2018revised_newcelltype, plasschaert2018single_newcelltype, luecken2019currentsctutorial_survey}. This high-level resolution, however, comes with a cost. scRNA-seq data are often extremely sparse and prone to various technical and biological noise, such as sequencing depth, batch effects, and cell-cycle phases \citep{svensson2018exponentialscaling, tanay2017scaling,luecken2019currentsctutorial_survey, hie2020computationalmethods_survey}. Various dimensionality reduction techniques have been developed to leverage intrinsic structures in the data \citep{heimberg2016lowdim} to map to a lower-dimensional latent space. These methods help facilitate downstream tasks like clustering and visualization, while avoiding the curse of dimensionality. Our work emphasizes probabilistic dimensionality reduction methods, which, through providing explicit probabilistic models for the data, allows for more interpretable models and uncertainty measures in the learned latent space.

In particular, we study a class of latent variable models known as Gaussian Process Latent Variable Models (GPLVMs) \citep{lawrence2004gaussian_ppca_gplvm}, which have recently been applied to scRNA-seq data \citep{campbell2015bayesian_scgplvm, buettner2015computational_scgplvm, ahmed2019grandprix_scgplvm, verma2020robust_scgplvm, lalchand2022modelling_scgplvm}. These models, which use Gaussian processes (GPs) to define nonlinear mappings from the latent space to data space, can incorporate prior information in the GP kernel function, motivating its use in single-cell transcriptomics data to model known or approximated covariate random effects, such as batch IDs and cell cycle phases. This approach is made scalable via mini-batching; however, the resulting Bayesian GPLVM model (BGPLVM) struggles to learn informative latent spaces for certain datasets \citep{lalchand2022modelling_scgplvm}.

In this work, we present an amortized BGPLVM  better fit to scRNA-seq data by leveraging design choices made in a leading probabilistic dimensionality reduction method called single cell variational inference (scVI) \citep{lopez2018deep_scvi}. While scVI has seen impressive performance in a variety of downstream tasks, it does not easily allow for interpretable incorporation of prior domain knowledge.

In Sections \ref{sec:background} and \ref{sec:model_mods}, we describe this model, providing a concise background on BGPLVMs and highlighting the model modifications. Section \ref{sec:results} then discusses (1) an ablation study demonstrating each components contribution to the model's performance via a synthetic dataset; (2) comparable performance to scVI for both the synthetic dataset and a real-world COVID-19 dataset \citep{stephenson2021single}; and (3) promising results for interpretably incorporating prior domain knowledge about cell-cycle phases in an innate immunity dataset \citep{kumasaka2021mapping_innateimmunity}. Our work shines a light on key considerations in developing a scalable, interpretable, and informative probabilistic dimensionality method for scRNA-seq data.

\section{Background}\label{sec:background}
This section provides a concise introduction to existing BGPLVM models from the literature.

\subsection{Amortized Stochastic Variational Bayesian GPLVM}\label{sec:bgplvm}
Given a training set comprised of $N$ $D$-dimensional observations $\Y = [\bm{y}_{1} \ldots \bm{y}_{N}]^{T} \in \R^{N \times D}$, we seek to represent our data with $Q$-dimensional embeddings $\X = [\bm{x}_{1}\ldots\bm{x}_{N}]^{T} \in \mathbb{R}^{N \times Q}$ which are latent and stochastic and $Q \ll D$ provides the dimensionality reduction. The probabilistic model describing the data can be written as follows:
\begin{align}
    \text{Latent prior: } p(\X) = \prod_{n = 1}^N \N(\x_n|\mathbf{0}, \I_Q) &\hspace{3mm}
   \text{GP Prior: } p(\F|\X, \theta) = \prod_{d = 1}^D \N(\f_d | \mathbf{0}, \K_{NN}),\\
  \text{Likelihood: } p(\Y|\F, \sigma_y^2) = \prod_{d = 1}^D \N(\y_d | \f_d,& \sigma_y^2 \I_N) = \prod_{n = 1}^N\prod_{d = 1}^D\N(y_{nd} | f_d(\x_{n}), \sigma_y^2),
\end{align}
$\F \equiv \{ \bm{f}_{d} \}_{d=1}^{D}$ denotes the collection of latent functions  where $\bm{f}_{d}$ is associated with $\bm{y}_{d}$ (the $d^{th}$ column of $\textbf{Y}$). $\K_{NN}$ is the covariance matrix corresponding to a user chosen positive-definite kernel function $k(\x, \x^{\prime})$ evaluated on latent points $\{\bm{x}_{n}\}_{n=1}^{N}$ and parameterized by hyperparameters $\bm{\theta}$. The kernel hyperparameters are shared across all dimensions $D$.

Moreover, to speed up computation and allow for mini-batching, we use inducing variables $\bU = \{\bu_m \in \R^Q\}_{m = 1}^M$ also distributed with a GP prior $\bu_d | \Z \sim \N (\0, \K_{MM})$, where $\K_{MM}$ is the kernel evaluated at inducing locations $\Z \in \R^{M \times Q}$ as in \citet{ hensman2013gaussian_svi_gp, lalchand2022generalised_gplvm}. The introduction of inducing variables gives us the following sparse GP prior:
\begin{align}
     %p(\U) &= \mathcal{N}(\bu_{d}|\0, \K_{MM})\\
     p(\F | \bU, \X) &= \prod_{d = 1}^D \N(\f_d | \K_{NM} \K_{MM}^{-1} \bu_d, \K_{NN} - \K_{NM}\K_{MM}^{-1}\K_{MN}).
\end{align}
The joint posterior over all unknowns $p(\F,\U,\X|\Y)$ is intractable, but admits a tractable lower bound to the marginal likelihood $p(\Y|\bm{\theta})$ under the variational formulation,
\begin{equation}
\begin{split}
 q(\F, \X, \U) = \Big[\prod_{d=1}^{D}p(\bm{f}_{d}|\bm{u}_{d},\X)q(\bm{u}_{d}) \Big]q(\X),
\label{titsias}
\end{split}
\end{equation}
where the variational distributions are:
\begin{align}
    % \prod_{n = 1}^N q(\x_{n}) = \prod_{n = 1}^N \N(\x_n | \bm{\mu}_n, \bm{\Sigma}_n)\\
    q(\X) =\prod_{n=1}^N \N(\x_n | H_{\phi_{1}}(\y_n), \text{diag}(H_{\phi_{2}} (\y_n)), \hspace{3mm}
    q(\U) = \prod_{d=1}^{D}q(\bu_d) = \prod_{d = 1}^{D} \N(\bu_d | \m_d, \mathbf{S}_d),
\end{align}
where $\{\m_d, \mathbf{S}_d\}_{d = 1}^D$ denotes the variational parameters. The mean and variance of the variational Gaussian distributions are parameterized as outputs of individual neural networks $H_{\phi_{1}}$ and $H_{\phi_{2}}$, which act as encoders. The network weights are amortized and shared across all the data points enabling extension to very large scale datasets as the amortized model side-steps the need to learn the latent variational means and covariances per data point, i.e. $q(\x_{n})$, and once trained, allows for constant-time $\mathcal{O}(N)$ inference.
% and  proximity in data space is preserved in the latent space. .
The resulting stochastic variational lower bound \citep{lalchand2022generalised_gplvm}, which factorizes across $N$ and $D$, permitting mini-batching, is given by:
    \begin{align}\label{elbo}
    \mathcal{L}(q(\cdot)) = \sum_{n, d}\mathbb{E}_{q(\bm{x}_{n})}\mathbb{E}_{p(\bm{f}_{d}|\bm{u}_{d}, \bm{x}_{n})q(\bm{u}_{d})}\left[\log \N (y_{nd} | f_d(\x_n), \sigma_y^2) \right] 
    - & \sum_{n=1}^N \KL\left(q(\x_n) || p(\x_n)\right) \\ - &\sum_{d=1}^D \KL\left(q (\bu_d) || p(\bu_d |\Z)\right).\nonumber
\end{align}

\subsection{Encoding Domain Knowledge through Kernels}\label{sec:background_kernels}
A key benefit of using GPLVMs is that we can encode prior information into the generative model, especially through the kernel design, allowing for more interpretable latent spaces and less training data. Here, we highlight kernels tailored to scRNA-seq data that correct for batch and cell-cycle nuisance factors as introduced by \citet{lalchand2022modelling_scgplvm}.

\paragraph{Batch correction kernel formulation}
In order to correct for confounding batch effects through the GP formulation, \citet{lalchand2022modelling_scgplvm} proposed the following kernel structure with an additive linear kernel term to capture random effects:
\begin{equation}\label{eq:aug_kernel}
      \tilde{\f}_d \sim \N(\mu_f \I_N + \Phi \zeta_d, \underbrace{\K_{NN}+ \nu \Phi \Phi^T)}_{\tilde{\K}_{NN}},
\end{equation}
which implicitly represents the relation $\Y = \F + \Phi\B + \mathbf{\varepsilon}$ where $\Phi \in \R^{N \times D_{\text{covar}}}$ is the design matrix where each row represents the known covariates for each cell; $\B \in \R^{D_{\text{covar}} \times D_{\text{gene}}}$ is a random variable $[\ldots, B_{d}, \ldots]$ ($B_{d}$ denotes a column) representing the random effect of each known covariate on gene expression $B_{d} \sim \mathcal{N}(\zeta_{d}, \nu\mathbf{I}_{D_{covar}})$, $\zeta_{d} \in \mathbb{R}^{D_{covar}}$, $\nu \in \mathbb{R}$, $\mathbf{\varepsilon} \in \R^{N \times D_{\text{gene}}}$ represents the noise model and $\mu_{f} \in \mathbb{R}$ is a constant mean for the latent functions. For most of this work, we use an SE-ARD kernel with additive linear kernel, henceforth denoted as SE-ARD+Linear.

\paragraph{Cell-cycle phase kernel}
When certain genes strongly reflect cell-cycle phase effects, obscuring key biological factors, a kernel designed to explicitly address a cell-cycle latent variable can effectively mitigate these effects. This motivates the use of adding a periodic kernel to the above kernel formulation. In particular, we specify the first latent dimension as a proxy for cell-cycle information and model our kernel as:
\begin{align}\label{eq:per_aug_kernel}
      k_{\tilde{f}} (\bm{x}, \bm{x'}) 
      &= \sigma_f^2 \exp\left(\frac{-2\sin^2(|\bm{x_1}-\bm{x_1'}|/2)}{l_1^2}\right) \times \exp\left(-\sum_{q = 1}^Q \frac{(\bm{x_q}-\bm{x_q'})^2}{2l_q^2}\right) + \nu \Phi \Phi ^T\\
      &= k_{\text{per}} \times k_{\text{se-ard}} + k_{\text{lin}},
\end{align}
where $\bm{\theta} = \left\{\sigma_f^2, \{l\}_{q = 1}^Q, \nu, \mu_f, \zeta_d\right\}$ are the hyperparameters of the BGPLVM. In particular, the periodic kernel helps capture the effects of the cell-cycle phases. We will refer to this kernel as PerSE-ARD+Linear, which will be used in our study of the innate immunity dataset discussed in Section \ref{sec:results}.

\section{Our model}\label{sec:model_mods}

\begin{figure}[h]
    \centering
    \includegraphics[scale= 0.35]{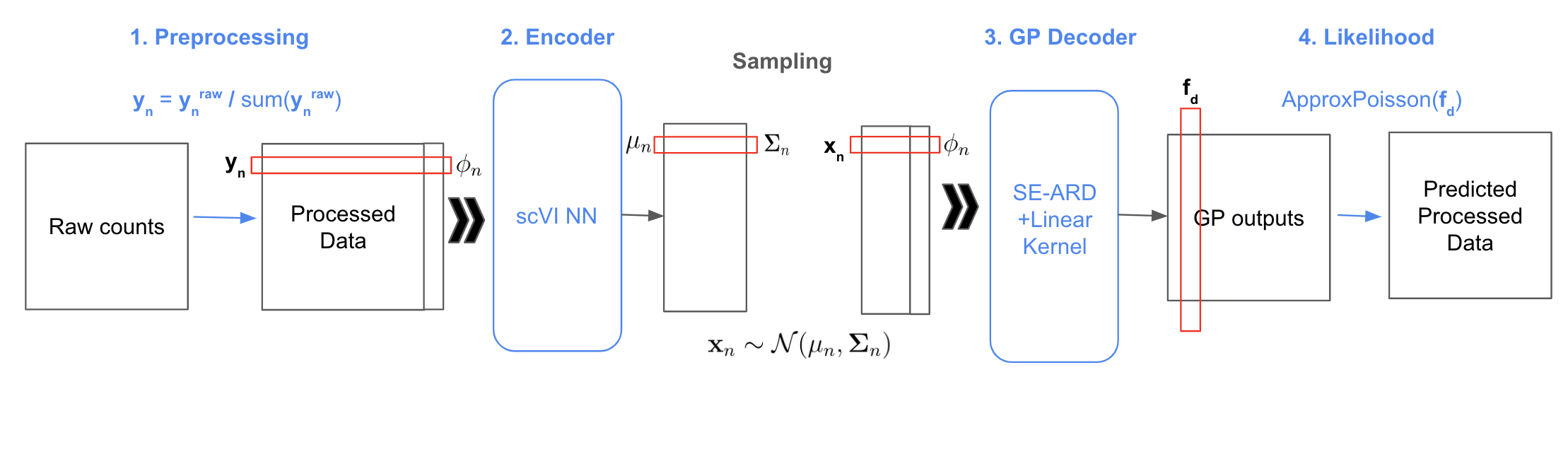}
    \vspace{-5mm}
    \caption{Overview of Modified BGPLVM Model}
    \label{fig:proposed_model_schematic}
\end{figure}

In the sections below, we discuss a set of modifications to the baseline model presented above, which form the main contributions of this work. In particular, we show that row (library) normalizing the data, using an appropriate likelihood, incorporating batch and cell-cycle information via SE-ARD+Linear and PerSE-ARD+Linear (Section \ref{sec:background_kernels}) and implementing a modified encoder significantly improves the BGPLVM's performance. We present the schematic of the modified BGPLVM in Figure \ref{fig:proposed_model_schematic}.

\subsection{Pre-Processing and Likelihood}\label{sec:design_likelihood}
Raw scRNA-seq data are discrete and must be pre-processed to better align with the Gaussian likelihood in the probabilistic model of the baseline discussed above, which we call OBGPLVM (short for Original Bayesian GPLVM). However, the assumption that this pre-processed data are normally distributed is not necessarily justified. Instead of adjusting the data to fit our model, we aim to better adapt our likelihood to the data. In particular, we only normalize the total counts per cell (i.e. library size) to account for technical factors \citep{normalizationlun2016} and adopt a negative binomial likelihood like that in scVI (detailed in Appendix \ref{app:baseline_models_scvi}).

In particular, we use a negative binomial with fixed scaling factor $\ell = 5000$ and $r = 10^6$. This is the simplest likelihood function and approximates a Poisson distribution. To account for sequencing depth differences, the likelihood requires the raw count data to be library size normalized to $5000$ first. We call this likelihood \textit{ApproxPoisson}. \footnote{The \href{https://docs.scvi-tools.org/en/stable/api/reference/scvi.distributions.NegativeBinomial.html}{scVI negative binomial} distribution's parameterization is equivalent to the generative model, $y|w \sim \text{Poisson}(w)$ with $w \sim \Gamma(\theta, \theta/\mu)$. Note that $w \overset{\theta \rightarrow \infty}{\sim} \mathcal{N}(\mu, \mu/\theta)
\rightarrow \delta(\mu)$, and thus $y \overset{\text{approx.}}{\sim} \text{Poisson}(\mu)$.}
    \begin{align}
        y_{nd} \sim \text{Negative Binomial}\left(5000 \times \text{softmax}(\tilde{f}_d(x_n)), 10^6\right).
    \end{align}

In our initial experiments, we found that the more complex the likelihood function was (in terms of parameters to be learned), the worse the resulting BGPLVM-learned latent space was. While one may expect the more complex and expressive likelihoods to perform better, this opposite trend may be because the model is non-identifiable. That is, especially since the loss function does not explicitly optimize for latent space representations, the extra parameters may overfit and cause the model to fail to learn important biological signals. One such ablation study is presented in Appendix \ref{app:mods_likelihood}. Due to this observation, we focus on the simplest (and best performing) negative binomial-based likelihood, \textit{ApproxPoisson}.

\subsection{Encoder}
In the encoder analysis, we compare a simple encoder comprised of linear layers followed by softplus activations (Simple NN) with the scVI's more complex encoder (scVI NN). scVI NN incorporates batch information as input to the nonlinear mapping, so incorporating this encoder into the BGPLVM may help address batch effects observed in the raw count data. Additionally, the scVI encoder architecture includes batch normalizations, contributing to a more stable optimization process, which we leverage for our GPLVM implementation. 

\section{Results and Discussion}\label{sec:results}
We present results for three experiments on an simulated dataset and two real-world datasets, which are detailed in Appendix \ref{app:data_experiment_details}. Full experiment details and results with latent space metrics are also presented in Appendix \ref{app:experiment_details} and \ref{app:expts_detailed_results}.

\subsection{Each component is crucial to modified model performance}
\begin{figure}[h]
    \centering
    \includegraphics[scale = 0.2]{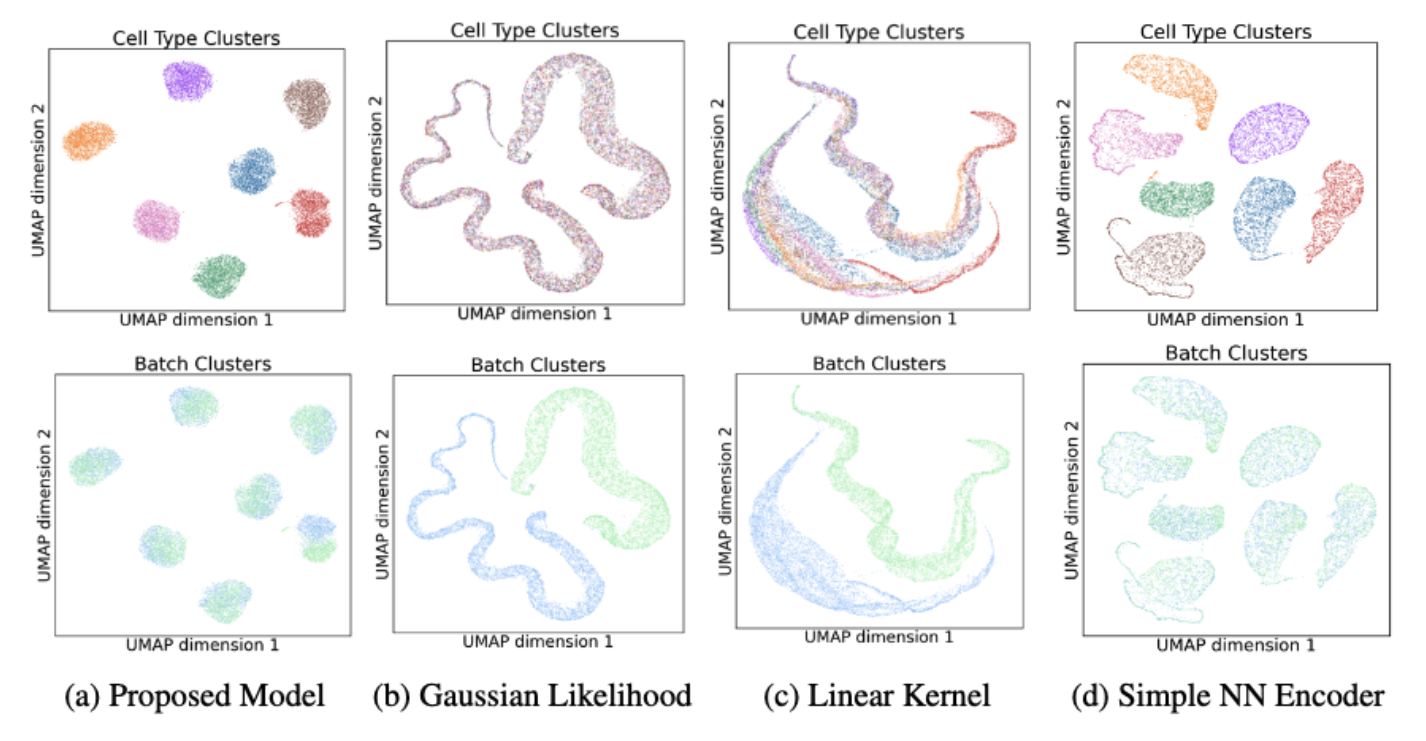}
    \caption{Ablation study with the simulated dataset on the proposed BGPLVM model where we change one component at a time (labeled in subfigures) and visualize the resulting UMAPs. The top row is colored by cell-type and the bottom row by batch.}
    \label{fig:ablation_study}
\end{figure}

To better understand how each component affects our model performance, we conducted an ablation study with a synthetic scRNA-seq dataset distributed according to a true negative binomial likelihood simulated by Splatter \citep{zappia2017splatter}. In particular, we reverted each component to a more standard BGPLVM component to evaluate its importance to the model's overall performance. The results for this experiment are detailed in Figure  \ref{fig:ablation_study} for the simulated dataset. 
Changing the pre-processing step and likelihood to match a Gaussian distribution as is done in standard GPLVMs completely removes any perceivable cell type separation and results in separated batches (Fig. \ref{fig:ablation_study}(b)). These observations support our hypothesis that the likelihoods were misaligned with the underlying distribution, at least for the simulated single-cell dataset. 

If the SE-ARD+Linear kernel is changed to a fully linear kernel (detailed in Appendix \ref{app:linear_kernel}), the batches separate while the cell-types begin to mix but are still slightly differentiable, albeit within the separated batches (Fig. \ref{fig:ablation_study}(c)). These changes may be attributed to the fact that linear kernel is not expressive enough to capture the cell-type information while the nonlinearity of the SE-ARD+Linear kernel permits extra flexibility.

In this reverse ablation study, the encoder exhibits the least impact on the latent space representation, as evidenced by the clear separation of cell types and well-mixed batches in Fig. \ref{fig:ablation_study}(d).  This behavior can be attributed to the encoder playing a smaller role in defining the generative model as it primarily functions as a means of regularization for mappings from the data space to the latent space.
\vspace{-6mm}

\subsection{Modified model achieves significant improvements over standard Bayesian GPLVM and is comparable to scVI} 

We compare our proposed model with three benchmark models: OBGPLVM, the current state-of-the-art scVI \citep{lopez2018deep_scvi} (Appendix \ref{app:baseline_models_scvi}), and a simplified scVI model with a linear decoder (LDVAE) \citep{svensson2020interpretable_linearscvi} (Appendix \ref{app:baseline_models_ldvae}) on the synthetic dataset and a real-world COVID-19 dataset \citep{stephenson2021single}. The UMAP plots for the COVID dataset are presented in Figure \ref{fig:covid_model_comparisons} and the detailed latent space metrics and UMAP plots are given in Appendix \ref{app:expts_detailed_results}.

Based on the UMAP visualizations, we observe that for both the simulated and COVID datasets, the modified BGPLVM achieves more visually separated cell types and mixed batches compared to the standard Bayesian GPLVM. The model also achieves visually comparable visualizations to scVI and LDVAE (Figures \ref{fig:simdata_model_comparisons} and \ref{fig:covid_model_comparisons}). While the modified model may not achieve better performance when compared to scVI and LDVAE, the GPLVM offers a more intuitive way to encode prior domain knowledge, and exploring such kernels and likelihoods more tailored to specific datasets are left for future work.

\begin{figure}[h]
    % \centering
    \includegraphics[scale=0.38]{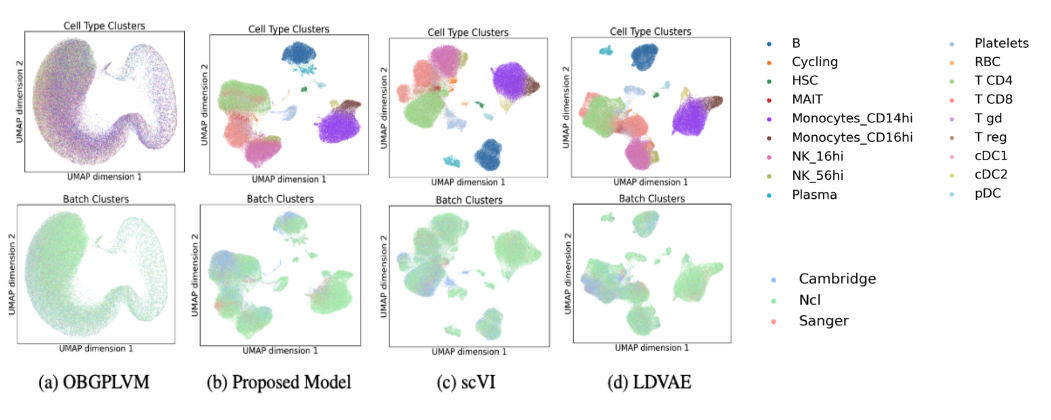}
    \caption{UMAPs generated from the latent spaces of four models: an implementation of the original BGPLVM, the modified BGPLVM for scRNA-seq data, scVI, and a linear decoder scVI (LDVAE) for the COVID data set. The top row is color/shaded by cell type and the bottom by batch.}
  \label{fig:covid_model_comparisons}
\end{figure}

\subsection{Consistency of Latent space with biological factors}
\begin{figure}[h]
    \centering
    \includegraphics[scale = 0.30]{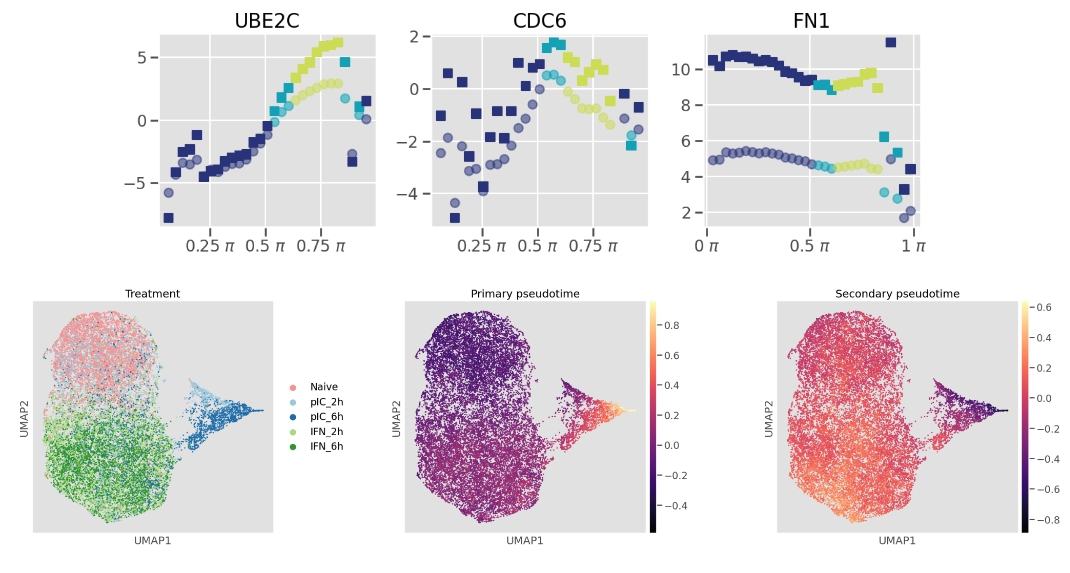}
    \caption{(Top row) Plots of log means and log variances (both parametrized by the same GP) versus learned cell-cycle pseudotime dimension for three specific genes (UBE2C, CDC6, FN1). The squares depict log variances and the circles depict log means of the library normalized data, both colored by the phases annotated in \citet{kumasaka2021mapping_innateimmunity}. We see that our model's learned cell-cycle phases correspond roughly to the phases labelled in \citet{kumasaka2021mapping_innateimmunity}. (Bottom row) UMAP plots of our model's learned latent space excluding directions identified with hidden technical effects (e.g. batch and plate border effects). Cells are colored by treatment condition (left), primary (middle) and secondary (right) pseudotime directions.}
    \label{fig:interpretability_expt}
\end{figure}

An advantage of our model is the ability to incorporate biologically interpretable data to boost latent space interpretability and overall performance. In particular, we compared our learned latent space with previous expert-labelled inferences on the innate immunity dataset in \citet{kumasaka2021mapping_innateimmunity}. Pretraining on well-initialized latents and finetuning our model with a PerSE-ARD+Linear kernel allowed us to recover latents consistent with those inferred and biologically motivated in \citet{kotliar2019identifying} (Figure \ref{fig:interpretability_expt} (top row)) while also separating cells by their treatment conditions (Figure \ref{fig:interpretability_expt} (bottom row)). Moreover, as indicated by the color gradations in the right two UMAP plots in the bottom row, the model's learned latent space is able to distinguish immune response pseudotime directions. This shows how initializations can be done on the amortized BGPLVM encoder-decoder models.

\section{Conclusion}
This paper identifies a misalignment in the generative model of current GPLVMs used in single-cell data and proposes an amortized BGPLVM better adapted to the scRNA-seq dimensionality reduction setting. In particular, by drawing insight from commonly used single-cell-specific methods, including scVI, LDVAE, and Splatter single-cell simulations, our proposed model tackles three main aspects of single-cell data by (1) accounting for count data with an approximate Poisson likelihood, (2) incorporating batch effect modelling in both the encoder and GP kernel, and (3) normalizing the library size in the data via a pre-processing step. We demonstrate the importance of aligning modelling choices to domain-specific knowledge as the model achieves comparable performance to scVI on both a simulated dataset and real-world COVID dataset in both UMAP visualizations and commonly used latent space metrics.


\subsubsection*{Acknowledgments}
The authors would like to thank Emma Dann, Natsuhiko Kumasaka and the rest of the team at Sanger for help and guidance with our initial project and for providing the data and code, which we based this study on. AR is supported by the accelerate programme for scientific discovery. During the time of this work, SZ was supported by the Churchill Scholarship.

\bibliography{mlgenx2024}
\bibliographystyle{mlgenx2024}

\appendix
\section{Baseline Models}\label{app:baseline_models}
\subsection{scVI}\label{app:baseline_models_scvi}
\begin{figure}[h] % current standin for scvi model
    \centering
    \includegraphics[scale=0.45]{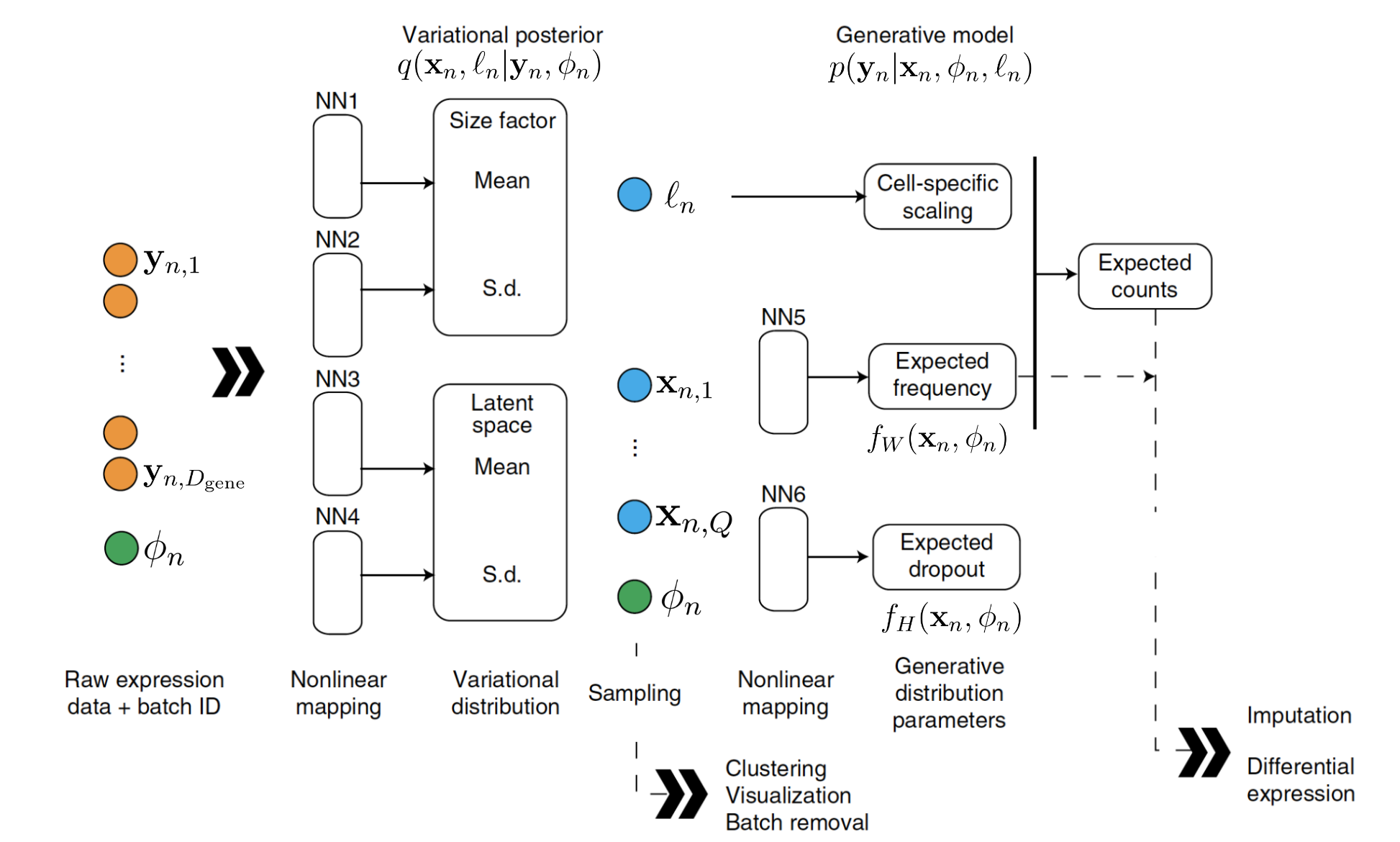}
    \caption{Overview of the scVI architecture adapted from  \citet{lopez2018deep_scvi}.}
    \label{fig:scvi_architecture}
\end{figure}

Proposed in 2019 by \citet{lopez2018deep_scvi}, single-cell variational inference (scVI) is a variational autoencoder that is tuned for single-cell data and has been shown to match current state of the art methods in a variety of downstream tasks, including clustering and differential expression \citep{lopez2018deep_scvi, luecken2022benchmarking_scib}. Furthermore, due to its neural network structure, the model is scalable to large datasets. An overview of the model is presented in Figure \ref{fig:scvi_architecture}.

We highlight several key components of the model that target phenomena commonly seen in single-cell data: (1) count data, (2) batch effect, and (3) library size normalization.

\textbf{Count Data.} As scRNA-seq raw count data are discrete, scVI adopts various discrete likelihoods, such as the negative binomial likelihood, for its models. This allows the model to learn a latent space directly from the raw expression data without any conventional pre-processing pipelines. Note that the original paper uses the zero-inflated negative binomial likelihood for the main model to account for dropouts, where gene expressions for a cell are not detected due to technical artifacts \citep{lopez2018deep_scvi, luecken2019currentsctutorial_survey}.

\textbf{Accounting for Batch Effects.} scVI also models for any effects from different sampling batches by incorporating batch ID information for each cell in both the encoding and decoding portions of the VAE model. While batch information is incorporated as input to the neural network encoder and decoders, it is unclear how exactly the batch effects are modelled.

\textbf{Library Size Normalization.} The third component scVI accounts for is the differences in total gene expression count per cell, or library size, of the data. In the raw count data, each cell has different total gene counts, which may affect comparisons between cells and impact downstream analysis \citep{hie2020computationalmethods_survey}. As this difference in library size, or sequencing depth, may be a result of technical noise, scVI chooses to model a scaling factor $\ell$ stand-in for library size. This latent variable is modelled as a log normal as done in \citet{zappia2017splatter} mappings from the raw counts and batch information to the  mean and variance learned by the neural network encoder. To avoid conflating the effects of the scaling factor and of biological effects in the data, a softmax is applied to the output of the decoder before being multiplied by the scaling factor to determine the negative binomial likelihood mean. 
% That is, we assume the mapping from the latent space to the dataspace determines the probabilities of the gene expression, and the scaling factor accounts for any differences in observed sequencing depth.

With these three key components in mind, scVI's generative model for a given data point $\y_n$ as follows:
\begin{align}
    \text{Prior on latents: } p(\x_n) &= \N(\x_n | 0, \I_Q)\\
    \text{Prior on scaling factor: } p(\mathbf{\ell}_n) &= \text{LogNormal}(\mathbf{\ell}_n| \ell_{\mu}, \ell_{\sigma^2})\\
    \text{Likelihood: } p(\y_n|\x_n, \ell_n, \bm{\phi}_n) &= \text{NegativeBinomial}\left(\y_n 
| \ell_n \text{softmax}\left(f_W(\x_n, \bm{\phi}_n) \right), r \right),
\end{align}
where $\bm{\phi}_n$ represents the batch information of cell $n$, $f_W(\x_n, \bm{\phi}_n)$ is a neural network decoder incorporating batch information, and $\ell_\mu$ and $\ell_{\sigma^2}$ are given by the empirical mean and variance of the log library size in the batch containing data point $n$. Here, the negative binomial is parameterized by the mean and inverse dispersion, so that the model has mean $\ell_n \text{softmax}\left(f_W(\x_n, \bm{\phi}_n) \right)$ and shape $r$. In this parameterization, when $r \rightarrow \infty$, this distribution approaches a Poisson distribution with mean equivalent to $\ell_n \text{softmax}\left(f_W(\x_n, \bm{\phi}_n)\right)$.

The corresponding loss term for each data point is given by
\begin{align}
    \cL(q(\x, \ell)) = \E_{q(\x, \ell | \y, \bm{\phi})}\left[{\log p(\y | \x, \ell, \bm{\phi})}\right] - \KL(q(\x|\y, \bm{\phi}) || p(\x)) - \KL(q(\ell|\y, \bm{\phi}) || p(\ell)),
\end{align}
where the parameters to be optimized are the weights of the neural network encoders and decoders as well as the inverse dispersion factor $r$ of the negative binomial likelihood. The way in which the loss can be decomposed into terms for datapoint allows the model to be trained with mini-batching \citep{hoffman2013stochastic}.

While scVI has been shown to perform well in a variety of downstream tasks \citep{lopez2018deep_scvi, luecken2022benchmarking_scib}, its complex architecture (as seen in Figure \ref{fig:scvi_architecture}) and opaque incorporation of known nuisance variables like batch effects make the model and its inferences difficult to interpret. 

\subsection{LDVAE}\label{app:baseline_models_ldvae}
In response to this lack of interpretability in the original scVI, \citet{svensson2020interpretable_linearscvi} proposed a linear version of scVI, where the neural network decoder is replaced with a linear mapping. In particular the LDVAE model is defined in the generative way as follows:
\begin{align}
    &\text{Prior on the latent variables: } p(\x_n) = \N(\x_n | 0, \I_Q),\\
    &\text{Prior on scaling factor: } p(\mathbf{\ell}_n) = \text{LogNormal}(\mathbf{\ell}_n| \ell_{\mu}, \ell_{\sigma^2}),\\
    &\text{Likelihood: } p(\y_n|\x_n, \ell_n) = \text{NegativeBinomial}\left(\y_n| \ell_n \text{softmax}\left(\x_n \W^T\right), r \right),
\end{align}
where $\W$ represents the linear mapping. Note that the mapping from latent space to data space is not completely linear as a nonlinearity is introduced in the softmax function. Moreover, Svensson et al. explored applying a BatchNorm layer to the linearly decoded parameters and found it matched or improved model performance in reconstruction error and learning the latent space in a mouse embryo development dataset \citep{svensson2020interpretable_linearscvi, cao2019single}. This BatchNorm layer is thus adopted in the LDVAE model, which further obscures a straightforward interpretation of the mapping defined by the decoder. 

Thus, while the LDVAE model allows for a more interpretable mapping from the latent space to the dimension space when compared to scVI, the use of a library size surrogate and a not clearly defined incorporation of batch information through NNs make both models less interpretable. 

\section{Experiment Details}\label{app:experiment_details}
\subsection{Data}\label{app:data_experiment_details}
We evaluate these models with two datasets: (1) a simulated dataset using the single-cell simulation framework Splatter \citep{zappia2017splatter} and (2) a COVID-19 dataset \citep{stephenson2021single}. 

\textbf{Simulated Data.} As the focus of our work is to dissect the assumptions made in single-cell data, we build our model based on a synthetic scRNA-seq dataset generated by the Splatter Splat scRNA-seq simulation \citep{zappia2017splatter}. The data are modelled off of a negative binomial distribution based on a hierarchical Gamma-Poisson model, where the parameters are drawn from the dataset \citep{kotliar2019identifying}. % add more commentary about the modeeling (?) 
The data are simulated with seven cell types and two batches, with 10000 cells in each batch and 10000 genes per cell. We then remove cells with fewer than 200 total gene expression counts and genes that are expressed in three or fewer cells. This results in a synthetic dataset having 16016 cells and 8819 genes.

\textbf{COVID-19 Data.} The COVID-19 dataset \citep{stephenson2021single} is a real world dataset comprised of gene expression counts obtained from peripheral blood mononuclear cells. This dataset includes samples from 107 patients exhibiting different degrees of COVID-19 severity, as well as samples from 23 healthy individuals. There are three main sampling locations – Sanger, Cambridge, and Newcastle – and the dataset also includes sample ID (143 batches total), where the sample IDs have unique codes for the sampling locations. There are 143 such sample IDs and 18 cell types considered. For this project, we take a subsample of this dataset that takes 100 000 cells and 5000 most variable genes as determined by Scanpy \citep{wolf2018scanpy}.

\textbf{Innate Immunity Data.} The innate immunity dataset of \citet{kumasaka2021mapping_innateimmunity} is comprised of 22,188 primary dermal fibroblasts from 68 donors who were either in the control group or were exposed to two stimulants to mimic innate immune response: (1) dsRNA Poly(I:C) for primary antiviral and inflammatory responses and (2) IFN-beta for secondary antiviral response. There were a total of 4999 genes and 7 latent dimensions (including cell-cycle latents).

\subsection{Experimental Set-Up}
For each of the experiments, we train the model with batch size 300, learning rate 0.05, and three different seeds: 0, 42, and 123. For the synthetic dataset, we train with 50 epochs and for the COVID-19 dataset, we use 15 epochs, which is sufficient for convergence for the corresponding datasets. The latent space dimension is set to $Q=10$ for all models. For evaluation, we use seed 1 for all UMAP visualizations, and the latent metrics are reported with the average and standard deviation (each up to two decimal digits) over the three training runs for each model. We use the CSD3 high-performance computers for model training. 

\subsection{Extra Modifications and Experiments}
% In contrast to the OBGPLVM implementation, which uses a periodic $\times$ SE-ARD + linear kernel, we focus on two different kernels that do not include a periodic component as not all of our datasets have approximated cell-cycle information. 
\subsubsection{Linear Kernel}\label{app:linear_kernel}
For the ablation study, we also consider a linear kernel that models the augmented latent space information 
    \begin{align}
        k_{\text{linear}}(\tilde{\bm{x}}, \tilde{\bm{x}}')= \nu \tilde{\bm{x}} \tilde{\bm{x}}'^T,
    \end{align}
    where  $\tilde{\bm{x}} = [\bm{x}^T \; \bm{\phi}^T]^T \in \R^{Q + D_{\text{covar}}}$ is the augmented latent variable that includes covariate information $\bm{\phi}$. $\nu$ is a variance parameter. 
    
    The corresponding augmented GP with linear mean and linear kernel is given by:
    \begin{align}
        p(\tilde{f}_d |\tilde{X}) = \mathcal{N}(\tilde{f}_d|\mu_f \mathbf{1}_{N} + \tilde{X}w_d, \tilde{K}_{NN}),
    \end{align}
    where $\mathbf{1}_{N} \in \mathbb{R}^N$ is a vector of 1s, $w_d \in \R^{Q}$ defines the linear mean, $\tilde{X} \in \mathbb{R}^{N \times (Q \times D_{\text{covar}})}$ is the matrix of latent variables $\tilde{X}$ augmented by the known covariates $\bm{\Phi}$, and $\tilde{K}_{NN} = k_{\text{linear}} (\tilde{X}, \tilde{X})$.
    
    % \item An SE-ARD kernel with linear terms, which we will refer to as SE-ARD+Linear. Similar to the OBGPLVM kernel, the SE-ARD kernel accounts for latent variable effects and the linear term corrects for random effects from known covariates like batch ID. The kernel is defined as 
    % \begin{align}
    %     k_{\text{seard\_linear}}(\tilde{\x}, \tilde{\x}')= \exp\left(-\sum_{q = 1}^Q \frac{(x_q - x_q')^2}{2l_q^2}\right) + \nu \bm{\phi}\bm{\phi}'^T,
    % \end{align}
    % where $x_q$ is the $q$th entry in $\x$, $\l_q$ determines the importance of latent space dimension $q$ for each $q \in [Q]$, and $\nu$ is a variance parameter. 
    
    % The corresponding augmented GP is given by:
    % \begin{align}
    %     p(\tilde{\f}_d |\tilde{\X}) = \N(\tilde{\f}_d |\mu_f \mathbf{1}_{N} + {\Phi}\zeta_d, \tilde{\K}_{NN}),
    % \end{align}
    % where $\mathbf{1}_{N} \in \R^N$ is a vector of 1s and $\zeta_d \in \R^{Q}$ defines the linear mean on the observed covariates $\Phi$. $\tilde{\K}_{NN} = k_{\text{seard\_linear}} (\tilde{X}, \tilde{X})$.

    % This kernel allow us to decode the latent space with smooth non-linear mappings while also incorporating random effects from known covariates.

\subsubsection{Likelihoods}\label{app:mods_likelihood}
\begin{figure}[h!]
  \centering
  \begin{subfigure}{0.3\textwidth}
    \centering
    \includegraphics[width=\linewidth]{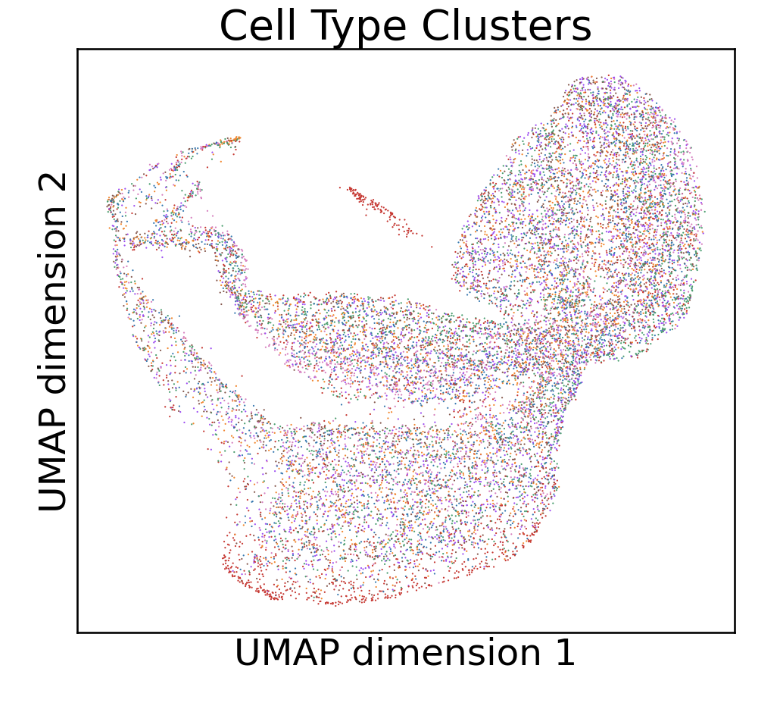}
  \end{subfigure}
   \begin{subfigure}{0.3\textwidth}
    \centering
    \includegraphics[width=\linewidth]{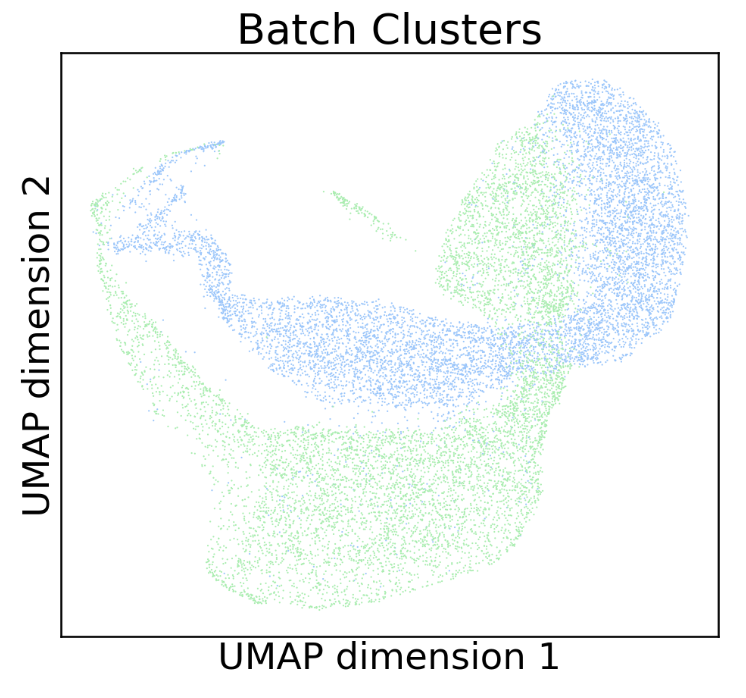}
  \end{subfigure}
  \caption{UMAP plots for an extended ablation study on the proposed model's likelihood. When library size is learned, the cell types become fully mixed (left) and the batches become separated (right).}
  \label{fig:ablation_study_likelihoods}
\end{figure}

In our ablation studies, more complex likelihoods (for example, a negative binomial likelihood where the library size of each row was learned) were observed to perform poorly, and likelihood simplifications like using the approximate Poisson likelihood led to improved performance (see Fig. \ref{fig:ablation_study_likelihoods}). This phenomenon could be explained by an issue with the identifiability of the model. The extra parameters in the model allow more flexibility in these likelihoods, but may also be learning and abstracting away pertinent cell-type information from the latent space variables. When the library size parameter is learned slowly, the model may be biased towards high-count cells, potentially disregarding the rest of the data and attributing latent space factors to technical noise rather than relevant biological differences. By constraining our likelihoods to slightly misaligned models, we may be encouraging the BGPLVM model to learn the 0s and smaller count values extremely well.

\section{Latent Space Metrics}\label{app:latent_space}
In this work, we compare these latent spaces both qualitatively and quantitatively. For qualitative measurements, we refer to UMAP 2-D visualizations \citep{mcinnes2018umap}. However, since UMAP is a stochastic mapping and the visualized distances between datapoints are not reflective of the true distances in the latent space \citep{mcinnes2018umap}, we also turn to quantitative latent space measurements pertinent to single-cell data. In particular, we focus on five quantitative metrics used in single-cell integration benchmarking that measure how well the latent space clusters cell types and how well the latent space mixes samples by batch \citep{luecken2022benchmarking_scib}. The measurements for cell-type separation in latent space are detailed in Bio Conservation Metrics (Section \ref{sec:scib_bio_metrics}) and measurements for batch mixing are detailed in Batch Correction Metrics (\ref{sec:scib_batch_metrics}).

In our experiments, following convention in \citet{luecken2022benchmarking_scib}, we average the batch variables to obtain an average batch metric score, and we average the bio-conservation metrics to obtain an average bio metric score. While \citet{luecken2022benchmarking_scib} propose an overall latent space score obtained through a weighted average of these two metrics, we deviate from this approach. We observed that models that failed to learn meaningful information could still yield high batch mixing scores (due to indiscriminate mixing) and consequently lead to misleading total scores. Hence, we choose to report only the average batch metric score and average bio metric score, separately.

% As an overview of the bio metrics and batch metrics, they are all normalized so that they mostly take on values between 0 and 1, with 0 indicating a `bad' latent space representation and 1 indicating a `good' latent space representation.

% Regarding clustering methods, we usually revert to k-means as it is interpretable and results should probably not be conflated with the clustering method.

\subsection{Bio Conservation Metrics} \label{sec:scib_bio_metrics}
We measure the latent space's ability to separate by cell-type with three different bio metrics: normalized mutual information (NMI), adjusted rand index (ARI), and cell-type average silhouette width (cellASW).

The NMI and ARI metrics require comparing cell-type information with learned clusterings from the latent space. To help make the metrics comparable for different models, we define the learned clusters with the Leiden clustering method with default resolution = 1 \citep{traag2019louvain_leiden_clustering} on the latent space projections. We also considered k-means clustering on the latent space projections but found that the resulting metrics were sometimes not reflective of the perceived clusters (e.g. when clearly-defined clusters are long and thin and close together width-wise, k-means outputs poor metrics).

\textbf{Normalized Mutual Information (NMI).} NMI compares the overlap of two clusterings, taking on values between 0 and 1 where 0 indicates no overlap and 1 indicates perfect overlap. 

More formally, let $T$ define the true cell type labels with $\#T$ distinct clusters and $C$ denote the predicted clusterings with $\#C$ distinct clusters. Furthermore, let $\{T_i\}_{i = 1}^{\#T}$ denote the different clusters in $T$ and $\{C_j\}_{j =1}^{\#C}$ denote the different clusters in $C$, and for each cluster $|C_j|$ is the number of samples in cluster $C_j$. $N$ is the total number of samples being clustered. Then, NMI is defined as follows:
\begin{equation}
NMI(T,P) = \frac{2I(T;C)}{H(T) + H(C)},
\end{equation}
where 
\begin{align}
   I(T; C) &= \sum_{i = 1}^{\#T} \sum_{j = 1}^{\#C} P(T_i \cap C_j) \log\left( \frac{ P(T_i \cap C_j) }{ P(T_i) P(C_j)}\right) \nonumber \\
   &= \sum_{i = 1}^{\#T} \sum_{j = 1}^{\#C}  \frac{|T_i \cap C_j|}{N} \log\left(\frac{N |T_i \cap C_j|}{|T_i| |C_j|}\right)
\end{align}
is the mutual information of $T$ and $C$ and 
\begin{align}
    H(T) = -\sum_{i = 1}^{\#T} \frac{|T_i|}{N} \log \frac{|T_i|}{N} 
\end{align}
denotes the entropy of $T$  (and is similarly defined for $C$).

% The scib module defaults to using Louvain clustering which builds off the idea of optimizing modularity (e.g. we want more edges within a cluster than between clusters). % TODO (?): discuss clustering methods more in depth?
% % TODO (?): talk about how Leiden is an improvement over Louvain.

% One main difference between the scib implementation \citep{luecken2022benchmarking_scib} and ours is that instead of optimizing the Leiden clustering resolution for the score, we choose a fixed arbitrary resolution so that we can compare across all our models.

\textbf{Adjusted Rand Index (ARI).} Adjusted Rand Index (ARI) also compares two clusterings but ARI (1) counts the pairwise agreements between the clusterings instead element-wise comparisons as done in NMI; and (2) adjusts for chance. The measurement usually takes on values between $0$ and $1$, and may extend to $-0.5$ for very different clusterings \citep{luecken2022benchmarking_scib}.

For a given sample $S$ of $N$ samples, Rand index by itself captures the proportion of samples upon which the two clusterings X and Y capture similar information. More formally,
\begin{align}
    RI = \frac{a + b}{a + b + c + d} = \frac{a + b}{\binom{N}{2}},
\end{align}
where
\begin{itemize}
    \item $a$ is the number of pairs that are in the same cluster in $T$ and in the same cluster in $C$
    \item $b$ is the number of pairs that are in different clusters in $T$ and in different clusters in $C$
    \item $c$ is the number of pairs that are in  same cluster in $T$ and in different clusters in $C$
    \item $d$ is the number of pairs that are in  different clusters in $T$ and in the same cluster in $C$
\end{itemize}

ARI is the corrected-for-chance version of RI.
\begin{align}
    ARI = \frac{\text{RI} - \text{Expected RI}}{\max(\text{RI}) - \text{Expected RI}}.
\end{align}

\textbf{Cell Average Silhouette Width (Cell type ASW).} 
The cell-type average silhouette width measures how compact the predicted clusters are by comparing the intra-cluster distances with inter-cluster distances. A score of 1 indicates well-separated and compact clusters while a score of 0 indicates misaligned or overlapping clusters. The clusters in this case are defined by the cell-types.

For a cell $n$ of cell type $C_j$, its silhouette score is defined as:
\begin{align}
    s(n) = \frac{b(n) - a(n)}{\max(a(n), b(n))},
\end{align}
where $a(n)$ is the average (Euclidean) distance between cell $n$ and the other cells of the same cell-type and $b(n)$ is the minimum average (Euclidean) distance between cell $n$ and a cell of a different cell type. More formally,
\begin{align}
    a(k) &= \frac{1}{|C_j| - 1} \sum_{l \in C_j} d(k,l), \\
    b(k) &= \min_{j' \neq j} \frac{1}{|C_{j'}|} \sum_{l \in C_{j'}} d(k, l).
\end{align}

Then, the average silhoutte width for each cell-type cluster $C_j$ is defined as the average silhoutte scores for each cell of that type:
\begin{align}
    \text{ASW}_{j} = \frac{1}{|C_j|} \sum_{n \in C_j} s(n).
\end{align}

Cell type ASW simply scales the average silhouette width over all cell-types so that instead of taking values between -1 and 1, it takes on values between 0 and 1:
\begin{align*}
    \text{cell type ASW}_j = \frac{1}{2} \left(1 + \frac{1}{M} \sum_{j = 1}^M ASW_j\right),
\end{align*}
where $M$ is the total number of cell types.

\subsection{Batch Correction Metrics}\label{sec:scib_batch_metrics}
\textbf{Batch Average Silhouette Width.} Much like Cell-type ASW, Batch ASW also measures the compactness of the predicted clusters. However, for the case of batches, we want the clusters to be spread out, so the Batch ASW formula must be adjusted accordingly so that a score of 1 reflects well-mixed batches and a score of 0 reflects poorly mixed batches. This is done by first introducing the absolute silhouette width for a cell $n$, 
\begin{align}
    s_{\text{batch}}(n) = |s(n)|,
\end{align} 
so that 0 represents a perfectly-mixed batch and any other value represents some deviation from being well-mixed.

The Batch ASW for a cell-type $j$ is then
\begin{align}
    \text{batch ASW}_j = \frac{1}{|C_j|} \sum{n \in C_j} 1 - s_{\text{batch}}(n).
\end{align}

The overall Batch ASW is given by
\begin{align}
    \text{batch ASW} = \frac{1}{M}\sum_{j = 1}^M \text{batch ASW}_j.
\end{align}

\textbf{Graph Connectivity}
The graph connectivity score represents how well the kNN graph connects cells of the same type. If there is good batch mixing, we would expect the cells of the same type to be clustered together, representing well connected same cell-type subgraphs. Conversely, when batches are not corrected for, cells of the same type could be dispersed across the latent space and not connected by the kNN graph. 

This idea is formally represented by the following graph connectivity metric:
\begin{align}
    \text{GC} = \frac{1}{M}\sum_{j=1}^M \frac{|\text{LCC}(G(N_j; E_j)}{|N_j|},
\end{align}
where $G(N_j; E_j)$ is the subgraph containing only cells of cell type $j$, $N_j$ is the set of nodes of cell-type $j$, and LCC$(G(N_j; E_j))$ is the Largest Connected Component of the subgraph $G(N_j; E_j)$.
\newpage
\section{Detailed Metrics}\label{app:expts_detailed_results}
We report the latent metrics for the first two experiments, taking the mean and standard deviation across trained models from three different seeds. Blue columns correspond to batch metrics and Green columns correspond to cell-type metrics.

\subsection{Ablation study}
\begin{table}[h]
\centering
\small
\begin{tabular}{|>{\raggedright\arraybackslash}m{1.8cm}|*{8}{>{\centering\arraybackslash}m{1.3cm}|}}
\hline
\rowcolor{pastelblue} \cellcolor{white}  \textbf{Model Change} &  \textbf{BatchASW} & \textbf{iLisi} & \textbf{kBET} & \textbf{Graph Connectivity} & \cellcolor{pastelgreen}\textbf{NMI Leiden} & \cellcolor{pastelgreen}\textbf{ARI Leiden} & \cellcolor{pastelgreen}\textbf{cellASW} &\cellcolor{pastelgreen} \textbf{cLisi} \\
\hline
\textbf{Simple NN} & 0.843 ± 0.003 & 0.881 ± 0.095 & 0.141 ± 0.245 & 0.338 ± 0.571 & 0.237 ± 0.394 & 0.110 ± 0.189 & 0.557 ± 0.157 & 0.511 ± 0.424 \\
\hline
\textbf{Gaussian Likelihood} & 0.447 ± 0.164 & 0.000 ± 0.000 & 0.015 ± 0.005 & 0.262 ± 0.158 & 0.003 ± 0.001 & 0.000 ± 0.000 & 0.429 ± 0.010 & 0.335 ± 0.010 \\
\hline
\textbf{Linear Kernel} & 0.088 ± 0.005 & 0.000 ± 0.000 & 0.000 ± 0.000 & 0.517 ± 0.001 & 0.627 ± 0.069 & 0.352 ± 0.100 & 0.486 ± 0.005 & 0.988 ± 0.005 \\
\hline
\end{tabular}
\caption{Latent space metrics for the ablation study on simulated dataset.}
\label{table:ablation}
\end{table}

\subsection{Benchmarking}
\begin{figure}[h]
  \centering
  \begin{subfigure}{0.22\textwidth}
    \includegraphics[width=\textwidth]{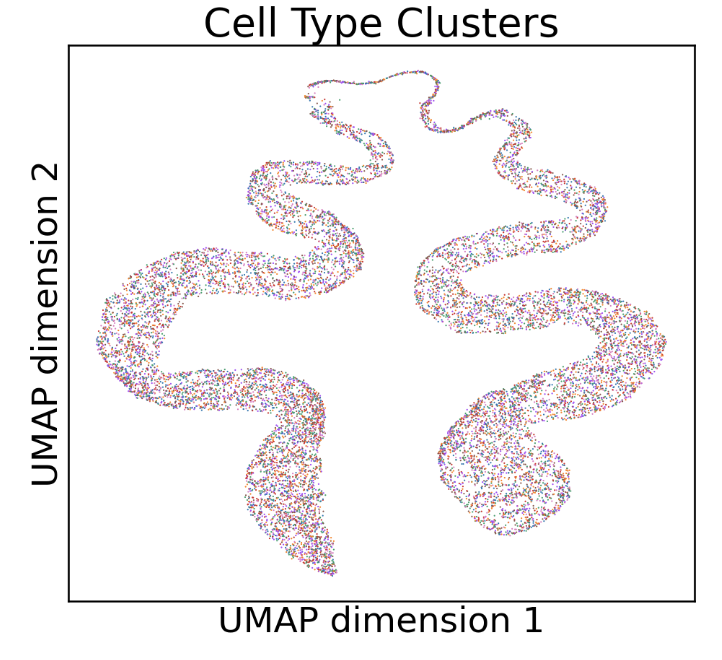}
  \end{subfigure}
  \begin{subfigure}{0.215\textwidth}
    \includegraphics[width=\textwidth]{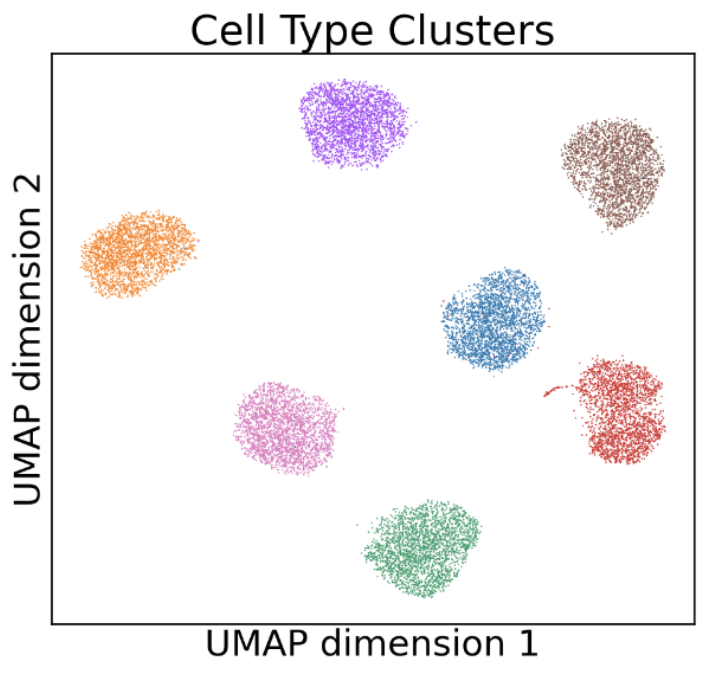}
  \end{subfigure}
  \begin{subfigure}{0.222\textwidth}
    \includegraphics[width=\textwidth]{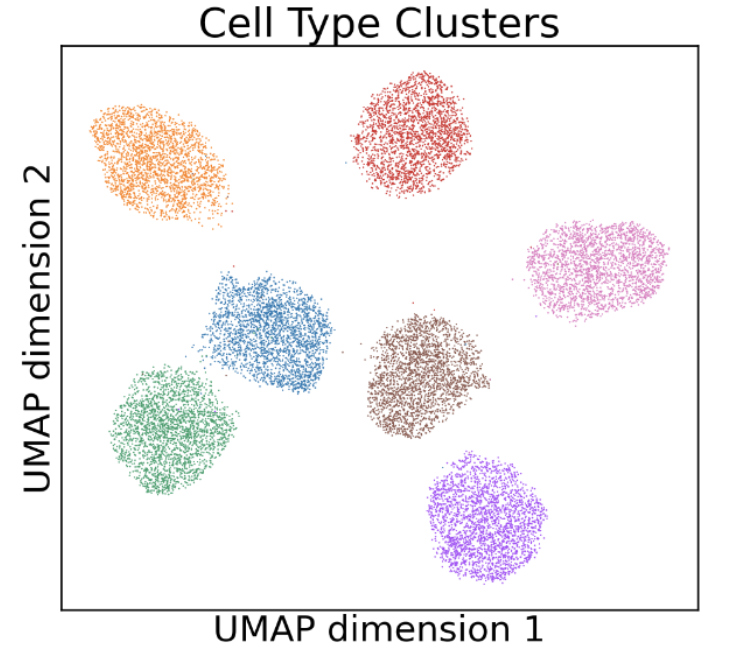}
  \end{subfigure}
  \begin{subfigure}{0.218\textwidth}
    \includegraphics[width=\textwidth]{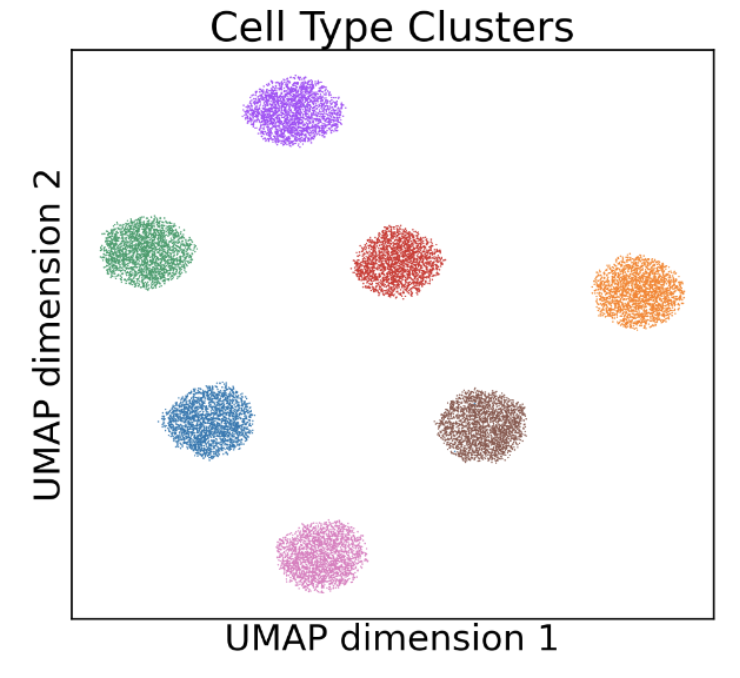}
  \end{subfigure}

  \begin{subfigure}{0.22\textwidth}
    \includegraphics[width=\textwidth]{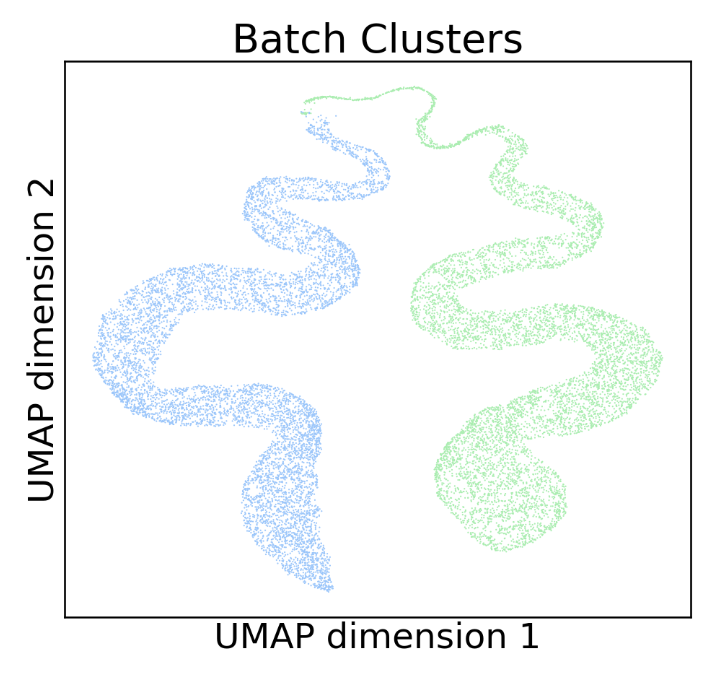}
    \caption{OBGPLVM}
  \end{subfigure}
  \begin{subfigure}{0.222\textwidth}
    \includegraphics[width=\textwidth]{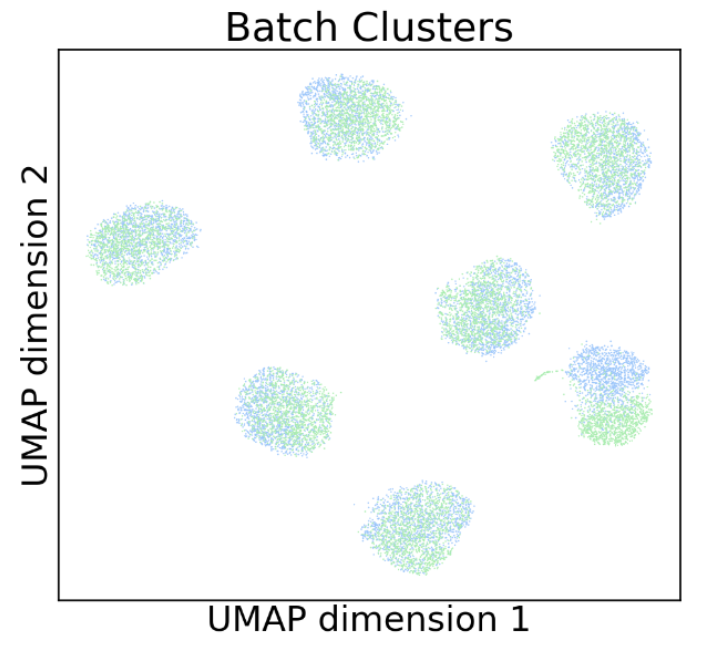}
    \caption{Proposed Model}
  \end{subfigure}
  \begin{subfigure}{0.22\textwidth}
    \includegraphics[width=\textwidth]{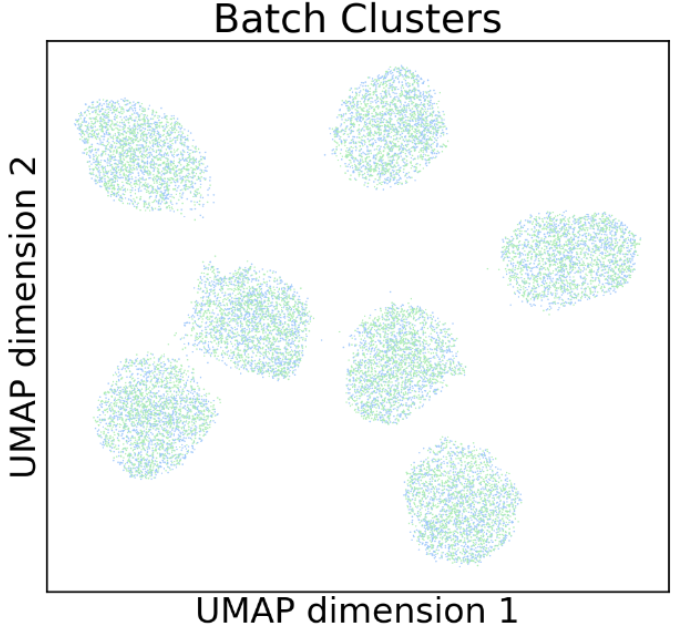}
    \caption{scVI}
  \end{subfigure}
  \begin{subfigure}{0.222\textwidth}
    \includegraphics[width=\textwidth]{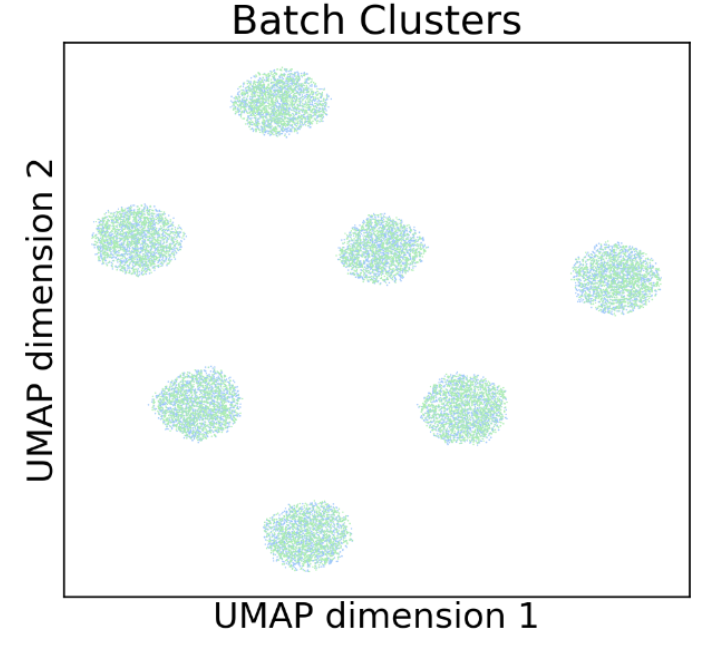}
    \caption{LDVAE}
  \end{subfigure}
  \caption{UMAPs generated from the latent spaces of four models: an implementation of the original BGPLVM, the modified BGPLVM for scRNA-seq data, scVI, and a linear decoder scVI (LDVAE) for the simulated dataset. The top row is color/shaded by cell type and the bottom by batch.}
  \label{fig:simdata_model_comparisons}
\end{figure}

\begin{table}[h]
\centering
\small
\begin{tabular}{|>{\raggedright\arraybackslash}m{1.8cm}|*{8}{>{\centering\arraybackslash}m{1.3cm}|}}
\hline
\rowcolor{pastelblue} \cellcolor{white}  \textbf{Model} &  \textbf{BatchASW} & \textbf{iLisi} & \textbf{kBET} & \textbf{Graph Connectivity} & \cellcolor{pastelgreen} \textbf{NMI} & \cellcolor{pastelgreen} \textbf{ARI} & \cellcolor{pastelgreen} \textbf{cellASW} & \cellcolor{pastelgreen} \textbf{cLisi} \\
\hline
\textbf{OBGPLVM} & 0.527 ± 0.472 & 0.319 ± 0.495 & 0.284 ± 0.424 & 0.582 ± 0.200 & 0.00247 ± 0.00036 & 0.00026 ± 0.00020 & 0.475 ± 0.004 & 0.321 ± 0.002 \\
\hline
\textbf{Proposed model} & 0.877 ± 0.049 & 0.570 ± 0.199 & 0.260 ± 0.149 & 0.998 ± 0.001 & 0.912 ± 0.061 & 0.849 ± 0.119 & 0.643 ± 0.026 & 1.0 ± 0 \\
\hline
\textbf{LDVAE} & 0.978 ± 0.049 & 0.913 ± 0.199 & 0.854 ± 0.149 & 1.000 ± 0.001 & 0.999 ± 0.061 & 1.000 ± 0.119 & 0.700 ± 0.026 & 1.0 ± 0.0 \\
\hline
\textbf{scVI} & 0.983 ± 0.002 & 0.916 ± 0.005  & 0.885 ± 0.028 & 1.000 ± 0.001 & 0.903 ± 0.061 & 0.805 ± 0.158 & 0.601 ± 0.012 & 1.000 ± 0.001 \\
\hline
\end{tabular}
\caption{Latent space metrics for benchmarking on the simulated dataset.}
\label{table:benchmark_simulated}
\end{table}

\begin{table}[h]
\centering
\small
\begin{tabular}{|>{\raggedright\arraybackslash}m{1.8cm}|*{8}{>{\centering\arraybackslash}m{1.3cm}|}}
\hline
\rowcolor{pastelblue} \cellcolor{white}  \textbf{Model} &  \textbf{BatchASW} & \textbf{iLisi} & \textbf{kBET} & \textbf{Graph Connectivity} & \cellcolor{pastelgreen} \textbf{NMI} & \cellcolor{pastelgreen} \textbf{ARI} & \cellcolor{pastelgreen} \textbf{cellASW} & \cellcolor{pastelgreen} \textbf{cLisi} \\
\hline
\textbf{OBGPLVM} & 0.795 ± 0.088 & 0.377 ± 0.055 & 0.465 ± 0.214 & 0.398 ± 0.258 & 0.265 ± 0.230 & 0.055 ± 0.065 & 0.459 ± 0.095 & 0.894 ± 0.086 \\
\hline
\textbf{Proposed model} & 0.848 ± 0.040 & 0.230 ± 0.017 & 0.884 ± 0.009 & 0.903 ± 0.028 & 0.606 ± 0.049 & 0.369 ± 0.053 & 0.492 ± 0.053 & 0.989 ± 0.002 \\
\hline
\textbf{linear scVI} & 0.918 ± 0.001 & 0.319 ± 0.003 & 0.861 ± 0.011 & 0.925 ± 0.001 & 0.690 ± 0.004 & 0.458 ± 0.007 & 0.578 ± 0.001 & 0.996 ± 0.000 \\
\hline
\textbf{scVI} & 0.945 ± 0.002 & 0.335 ± 0.004 & 0.881 ± 0.005 & 0.947 ± 0.002 & 0.722 ± 0.019 & 0.538 ± 0.053 & 0.544 ± 0.006 & 0.995 ± 0.000 \\
\hline
\end{tabular}
\caption{Latent space metrics for benchmarking on the COVID-19 dataset}
\label{table:benchmark_covid}
\end{table}

\end{document}